\documentclass[letterpaper]{article}
\usepackage{aaai2026}
\usepackage{times}
\usepackage{helvet}
\usepackage{courier}
\usepackage[hyphens]{url}
\usepackage{graphicx}
\usepackage{natbib}
\usepackage{caption}
\usepackage{algorithm}
\usepackage{algorithmic}
\usepackage{amsmath}
\usepackage{amssymb}

\title{Data Whitening Improves Sparse Autoencoder Learning}

\author{
    Ashwin Sarawatula,
    David Klindt
}
\affiliations{
    Case Western Reserve University, 
    Cold Spring Harbor Laboratory
}

\begin{document}

\maketitle

\begin{abstract}
Sparse autoencoders (SAEs) have emerged as a promising approach for learning interpretable features from neural network activations. However, the optimization landscape for SAE training can be challenging due to correlations in the input data. We demonstrate that applying PCA Whitening to input activations---a standard preprocessing technique in classical sparse coding---improves SAE performance across multiple metrics. Through theoretical analysis and simulation, we show that whitening transforms the optimization landscape, making it more convex and easier to navigate. We evaluate both ReLU and Top-K SAEs across diverse model architectures, widths, and sparsity regimes. Empirical evaluation on SAEBench, a comprehensive benchmark for sparse autoencoders, reveals that whitening consistently improves interpretability metrics, including sparse probing accuracy and feature disentanglement, despite minor drops in reconstruction quality. Our results challenge the assumption that interpretability aligns with an optimal sparsity--fidelity trade-off and suggest that whitening should be considered as a default preprocessing step for SAE training, particularly when interpretability is prioritized over perfect reconstruction.
\end{abstract}

\section{Introduction}

Sparse autoencoders (SAEs) \citep{ng2011sparse} have become a cornerstone of mechanistic interpretability, enabling researchers to extract human-understandable features from the internal activations of large language models (LLMs) \citep{bricken2023monosemanticity,cunningham2023sparse}. Individual neurons are often polysemantic, encoding multiple unrelated concepts simultaneously \citep{elhage2022toy}, making it difficult to isolate meaningful representations. By learning sparse, overcomplete dictionaries, SAEs decompose neural activations into latent dimensions that align with meaningful concepts or functions within a model’s computation \citep{marks2024sparse, kharlapenko2025scaling,klindt2025superposition}.

Despite their promise, training SAEs remains challenging. The optimization landscape is complex \citep{evci2020difficulty}, and finding features that are both interpretable and faithful to the original representations requires careful tuning of sparsity penalties and architecture choices \citep{gao2024scaling, templeton2024scaling,bussmann2025matryoshka,o2024compute}---but these methods still operate on correlated data, leaving the structure of the activation space unchanged.

\begin{figure}[h!]
    \centering
    \includegraphics[width=0.99\linewidth]{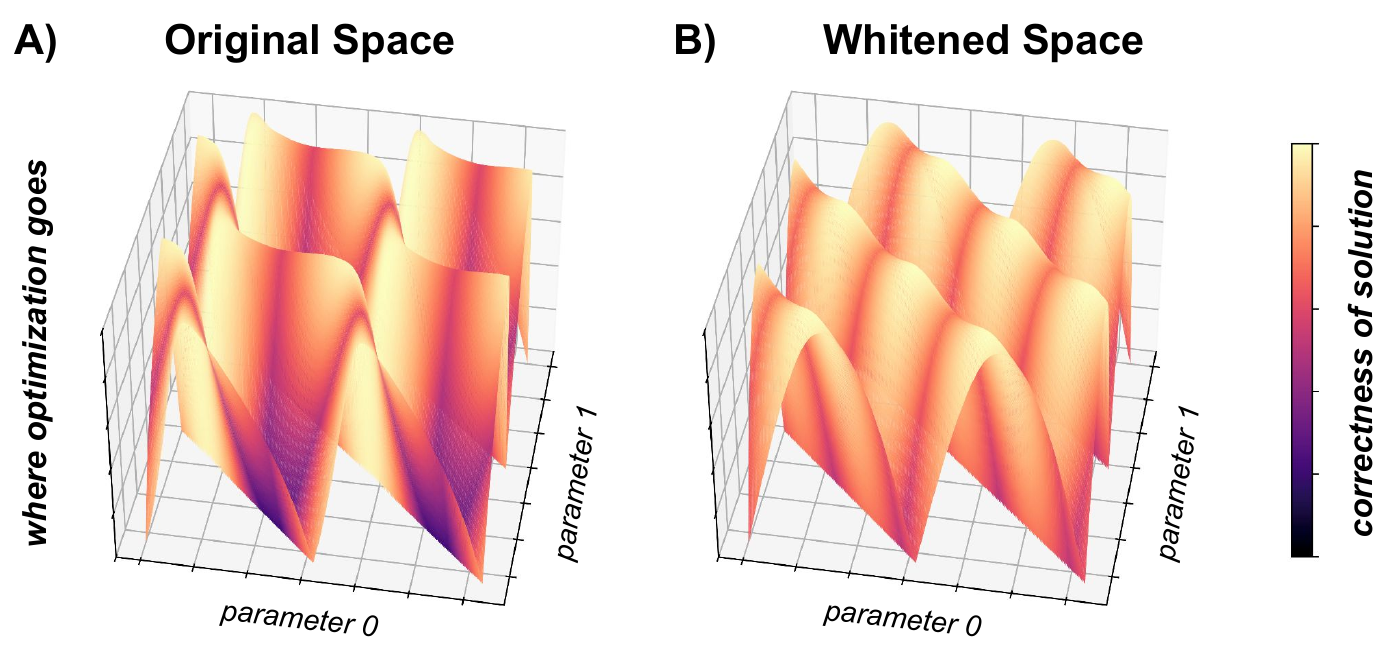}
    \caption{\textbf{Whitening transforms the optimization landscape.} 
3D visualization of the sparse coding landscape over all dictionary angles $(\theta_0, \theta_1) \in [0, 2\pi]^2$. 
Surface height shows sparsity (higher = sparser); color indicates feature recovery quality (brighter = better). 
\textbf{A}: Without whitening, high sparsity regions (peaks) are misaligned with accurate feature recovery (bright). 
\textbf{B}: After whitening, the landscape becomes isotropic and sparsity aligns with feature quality.}
\label{fig:theory_landscape}
\end{figure}

We revisit a classical idea from sparse coding and neuroscience: \textit{data whitening} via principal component analysis (PCA). While whitening is standard practice in classical sparse coding algorithms \citep{olshausen1996emergence, lee2006efficient} and independent component analysis (ICA) \citep{hyvarinen2001independent}, it has been largely overlooked in modern SAE training. Whitening removes correlations and equalizes variance in the input space, simplifying optimization and encouraging more independent features. In the brain’s visual system, whitening is believed to occur in the retina, where early sensory circuits decorrelate visual inputs to improve feature separability \citep{atick1990towards,atick1992does,olshausen1997sparse,graham2006can}. Motivated by this, we apply PCA Whitening as a preprocessing step for SAE training. This simple modification yields substantial gains in interpretability without altering model architecture or loss design. 
Our key contributions are:

\begin{itemize}
\item We provide theoretical analysis showing how whitening improves the SAE optimization landscape, making it more convex and isotropic.
\item We present simulation studies demonstrating the practical benefits of whitening for optimization.
\item We conduct comprehensive experiments on SAE Bench showing that whitening improves interpretability metrics including sparse probing accuracy (+7.3\%), spurious correlation removal (+54.03\%), and targeted probe perturbation (+372\%).
\end{itemize}

\section{Related Work}

\paragraph{Sparse Autoencoders.} 
Recent work has explored improving the interpretability and faithfulness of SAEs. Top-K SAEs \citep{gao2024scaling} fix the number of active latents per input, removing the need for an $L_1$ penalty. BatchTop-K \citep{bussmann2024batchtopk} extends this to the batch level to enhance reconstruction, while Gated and JumpReLU SAEs \citep{rajamanoharan2024Gated,rajamanoharan2024JumpRelU} address shrinkage and fidelity issues. Matryoshka SAEs \citep{bussmann2025matryoshka} enable hierarchical feature discovery, and \citet{o2024compute} find that SAEs do not fully solve sparse coding, though deeper encoders may help. Our work complements these methods by showing that PCA whitening improves interpretability without altering model architecture or loss. 

\paragraph{Whitening in Unsupervised Learning.}
Whitening has been standard in classical sparse coding since Olshausen and Field's seminal work \citep{olshausen1996emergence}. Efficient coding theory \citep{barlow1961possible} provides theoretical justification for whitening as optimal preprocessing \citep{atick1990towards,atick1992does}. In ICA, whitening is standard preprocessing because it simplifies the problem of demixing latent variables, which are assumed to be white \citep{hyvarinen2001independent}.
Its application to modern deep learning has been explored in some contexts. 
\citet{coates2011analysis} examined whitening as preprocessing for single-layer networks 
in unsupervised feature learning, finding improvements in classification tasks. 
However, whitening has been largely overlooked in the context of modern SAE training 
for mechanistic interpretability of LLMs. Our work provides the first systematic 
evaluation of whitening for SAEs on the SAEBench benchmark, demonstrating 
substantial improvements in interpretability metrics beyond reconstruction quality.

\paragraph{Interpretability Evaluation.} Recent work has established standardized frameworks for quantifying interpretability \citep{leavitt2020towards}. \citet{karvonen2025saebench} introduced \textit{SAEBench}, a large-scale benchmark that systematically compares SAE architectures across diverse interpretability metrics. \citet{karvonen2024measuring} proposed domain-grounded evaluation using structured tasks such as Chess and Othello, while \citet{kantamneni2025sparseautoencodersusefulcase} evaluated SAEs on real-world downstream probing tasks, highlighting limitations in current interpretability methods.
Visual and non-linguistic interpretability metrics are proposed and discussed in \citep{zimmermann2021well,klindt2023identifying,zimmermann2024measuring,klindt2025superposition,paulo2025evaluating}.

\section{Background}

\subsection{Sparse Autoencoders}

A sparse autoencoder learns to decompose neural network activations $\mathbf{x} \in \mathbb{R}^d$ into a sparse, overcomplete representation. The model consists of an encoder that maps inputs to features $\mathbf{f} = \sigma(\mathbf{W}_e \mathbf{x} + \mathbf{b}_e) \in \mathbb{R}^m$ where $m > d$, and a decoder that reconstructs the input as $\hat{\mathbf{x}} = \mathbf{W}_d \mathbf{f} + \mathbf{b}_d$.
The training objective minimizes reconstruction error while promoting sparsity:
\begin{equation}
\mathcal{L}(x) = \|x - \hat{x}\|_2^2 + \lambda S(f),
\end{equation}
where $S(f)$ penalizes dense activations and $\lambda$ controls the sparsity–reconstruction trade-off.

\subsubsection{ReLU SAEs.}  
ReLU-based SAEs impose \emph{soft sparsity} by adding an $L_1$ penalty to the latent activations:
\begin{equation}
\mathcal{L}_{\text{ReLU}} = \|x - \hat{x}\|_2^2 + \lambda \|\mathbf{f}\|_1
\end{equation}
This encourages gradual sparsity, allowing some activations to vary smoothly while penalizing dense features.

\subsubsection{Top-K SAEs.}  
Top-K SAEs instead enforce \emph{hard sparsity} by keeping only the $k$ largest activations per input:
\begin{equation}
\mathbf{f} = \mathrm{TopK}(\mathbf{W}_e \mathbf{x} + \mathbf{b}_e, k)
\end{equation}

\subsection{Classical Sparse Coding and Whitening}

In classical sparse coding \citep{olshausen1996emergence}, data whitening is a standard preprocessing step. Given data $\mathbf{x}$ with covariance $\mathbf{C} = \mathbb{E}[\mathbf{x}\mathbf{x}^\top]$, whitening transforms the data to $\mathbf{z} = \mathbf{C}^{-1/2}\mathbf{x}$, ensuring the covariance of $\mathbf{z}$ is the identity matrix.

This preprocessing provides several important benefits for model training. First, whitening makes the optimization landscape more isotropic, preventing bias toward high-variance directions. Second, whitening removes correlations in the input data that interfere with feature learning, allowing the model to discover more independent and disentangled representations. Third, whitening improves conditioning of the reconstruction problem by equalizing variance across dimensions, which stabilizes gradient updates and accelerates optimization. Despite these benefits, whitening has been largely underutilized in modern SAE training.

\subsection{SAEBench Metrics}
SAEBench is a comprehensive evaluation suite designed to assess  SAEs across multiple interpretability dimensions. 
We evaluate SAEs across five key SAEBench metrics:

\paragraph{CE Loss Score:} Measures how well SAE-reconstructed activations preserve the model’s next-token prediction accuracy. Higher scores indicate more faithful reconstruction.

\paragraph{Explained Variance:} Fraction of variance in the original activations preserved by SAE reconstruction. Higher values indicate better reconstruction fidelity.

\paragraph{Sparse Probing (Top 1):} Accuracy of linear probes trained on SAE activations across 35 binary diagnostic tasks. Higher values indicate more localized and interpretable features.

\paragraph{SCR (Top 20):} Evaluates an SAE’s ability to remove spurious correlations by ablating features associated with confounding variables(e.g., \textit{gender} and \textit{profession}). Higher values reflect stronger disentanglement and debiasing.

\paragraph{TPP (Top 20):} Measures causal specificity by ablating latents linked to a target class and assessing selective performance drop across all classes. Higher scores indicate more disentangled and causally isolated features.

\subsubsection{Metric Selection}
We selected these five metrics as they best capture reconstruction quality and latent interpretability, while remaining robust for models at the parameter scale ($<2$B) used in our experiments \citep[see note:][]{karvonen2025github}.


\section{Theoretical Analysis}

\subsection{How Whitening Reshapes the Optimization Landscape}

We provide a theoretical analysis demonstrating why whitening improves SAE training. Consider a simple 2D sparse coding problem where observations $\mathbf{y} \in \mathbb{R}^2$ are generated by mixing sparse sources $\mathbf{z} \in \mathbb{R}^2$ through a dictionary $\mathbf{A} \in \mathbb{R}^{2 \times 2}$:
\begin{equation}
\mathbf{y} = \mathbf{A}\mathbf{z}
\end{equation}

The goal of sparse coding is to recover the true dictionary $\mathbf{A}$ by learning an inverse mapping $\mathbf{W}$ such that the reconstructed sources $\hat{\mathbf{z}} = \mathbf{W}\mathbf{y}$ are sparse. The optimization landscape depends on two competing objectives:
i) \textbf{Sparsity}: The learned features should be sparse, measured by the inverse of the mean L1 norm: $\mathcal{S}(\mathbf{W}) = \frac{1}{\mathbb{E}[|\mathbf{W}\mathbf{y}|]}$
ii) \textbf{Feature Recovery}: The learned dictionary should align with the true dictionary, measured by the \textit{mean correlation coefficient} \citep[][see Appendix~\ref{appendix:simulation}, Eq.~\eqref{eq:mcc}]{hyvarinen2017nonlinear}.

An ideal optimization landscape should exhibit high values for both metrics at the same location in parameter space, with a smooth, convex basin leading to this optimum.

\subsection{Simulation: Optimization Landscape Analysis}

\begin{figure}[h]
    \centering
    \includegraphics[width=0.99\linewidth]{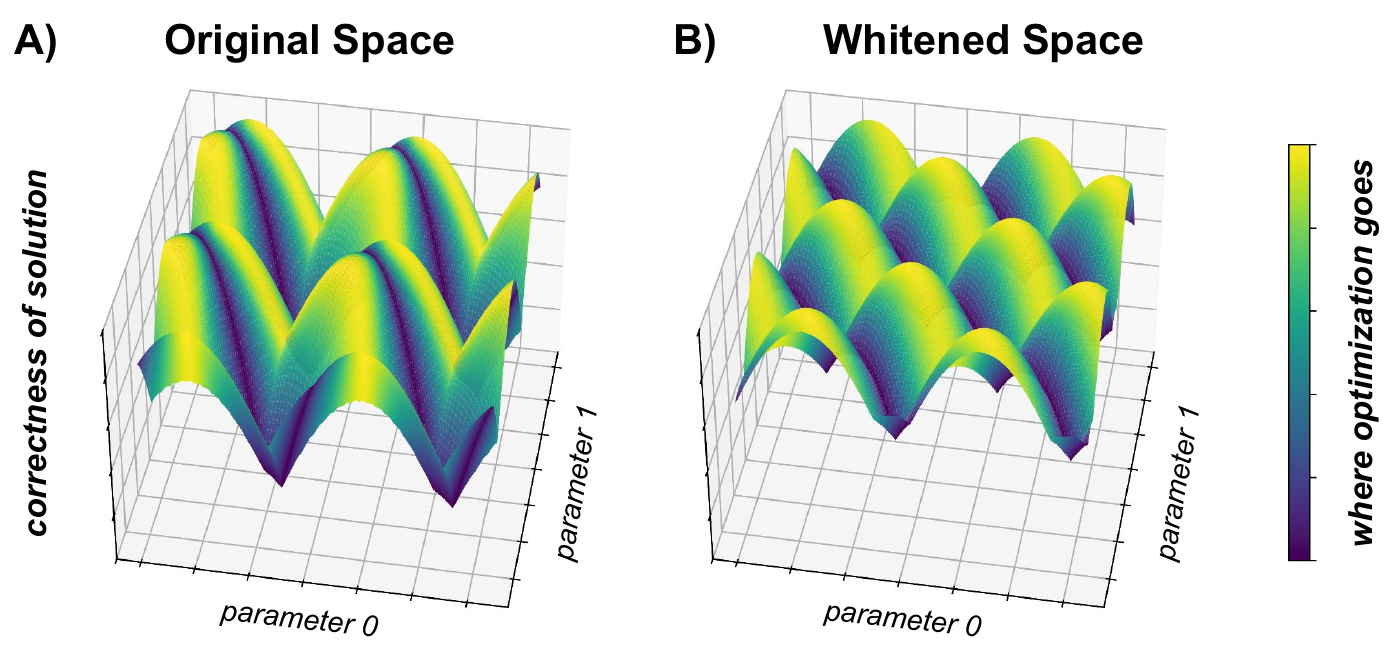}
    \caption{\textbf{Complementary view of optimization landscape.} 
Surface height shows feature recovery quality; color indicates sparsity level (brighter = sparser). 
\textbf{A}: Optimizing for sparsity (climbing to bright) may lead to poor feature recovery. 
\textbf{B}: After whitening, pursuing sparsity naturally yields interpretable features (bright colors at peaks).}
\label{fig:theory_landscape_flipped}
\vspace{-10pt}
\end{figure}

To visualize how whitening affects these landscapes, we conducted a systematic simulation study. We generated synthetic 2D data with correlated features by mixing sparse sources through a random dictionary. We then computed both the sparsity metric $\mathcal{S}(\mathbf{W})$ and feature recovery metric $\mathcal{R}(\mathbf{W}, \mathbf{A})$ across a dense grid of possible dictionary angles $(\theta_0, \theta_1) \in [0, 2\pi]^2$, where:
\begin{equation}
\mathbf{W}(\theta_0, \theta_1) = \begin{bmatrix} \cos\theta_0 & \sin\theta_0 \\ \cos\theta_1 & \sin\theta_1 \end{bmatrix}
\end{equation}

This exhaustive search over the parameter space reveals the complete structure of the optimization landscape.

\paragraph{Figure Interpretation.} Figure~\ref{fig:theory_landscape} shows how whitening transforms the optimization landscape by comparing before (A) and after (B) whitening. The surface height represents sparsity level (higher = sparser) while colors indicate feature recovery quality (brighter = better recovery). 

Without whitening (A), the landscape exhibits a narrow, elongated basin with steep gradients and poor conditioning. Critically, the highest sparsity regions (tall peaks) are misaligned with accurate feature recovery (which appears in duller colors), meaning that achieving high sparsity does not guarantee interpretable features. After whitening (B), the landscape becomes more isotropic with a wider basin. The bright colors now concentrate at the peaks, showing that sparsity and feature recovery are aligned—the sparsest solutions also yield the best feature reconstruction.

Figure~\ref{fig:theory_landscape_flipped} provides the complementary perspective, with surface height representing feature recovery quality and colors indicating sparsity (brighter = sparser). This view reveals that without whitening (A), pursuing bright regions (high sparsity) can lead to valleys rather than peaks (poor feature recovery). After whitening (B), the brightest colors align with the highest peaks, confirming that optimizing for sparsity naturally leads to better feature recovery.

\subsection{Implications for SAE Training}

These geometric transformations have direct consequences for SAE training. Whitening equalizes the eigenspectrum of the data covariance, transforming ill-conditioned problems into well-conditioned ones and leading to more stable gradient updates. In correlated data, sparsity and feature interpretability can be misaligned—pursuing sparsity may not yield semantically meaningful features. Whitening aligns these objectives, ensuring that sparse solutions correspond to interpretable features. While sparse coding is inherently non-convex, whitening makes the landscape more convex-like with a smoother basin around the global optimum, reducing sensitivity to initialization and hyperparameters. Finally, whitening enforces second-order independence by removing correlations, providing an inductive bias that encourages feature disentanglement.
Detailed simulation methodology is provided in Appendix~\ref{appendix:simulation}.
While our 2D simulation provides intuition for these effects, quantitative analysis of 
conditioning measures and gradient dynamics on high-dimensional real activations remains 
an important direction for future theoretical work.

\section{Experiments}

\begin{table*}[t]
\centering
\begin{tabular}{l|cc|c|c|c}
\hline
Metric & ReLU & +Whitening & $\Delta$ & \% $\Delta$ & $p$-value \\
\hline
CE Loss Score & \textbf{0.980} $\pm$ 0.005 & 0.954 $\pm$ 0.006 & -0.026 & -2.64\% & \(2.86 \cdot 10^{-5}\) \\
Explained Variance & \textbf{0.813} $\pm$ 0.028 & 0.772 $\pm$ 0.027 & -0.041 & -5.02\% & \(2.84 \cdot 10^{-6}\) \\
Sparse Probing (Top 1) & 0.757 $\pm$ 0.008 & \textbf{0.812} $\pm$ 0.008 & +0.054 & +7.15\% & \(1.05 \cdot 10^{-5}\) \\
SCR (Top 20) & 0.176 $\pm$ 0.015 & \textbf{0.271} $\pm$ 0.019 & +0.095 & +54.03\% & \(3.25 \cdot 10^{-6}\) \\
TPP (Top 20) & 0.021 $\pm$ 0.004 & \textbf{0.098} $\pm$ 0.010 & +0.078 & +372.00\% & \(5.66 \cdot 10^{-6}\) \\
\hline
\end{tabular}
\caption{ReLU architecture: SAE performance metrics averaged across all configurations (both models, all widths, all sparsity penalties). Values shown as mean $\pm$ SEM. $\Delta$ represents the difference between whitened and standard SAE. Bold indicates significantly better performance ($p < 0.05$).}
\label{tab:relu_results1}
\end{table*}

\begin{figure*}
    \centering
    \includegraphics[width=0.99\linewidth]{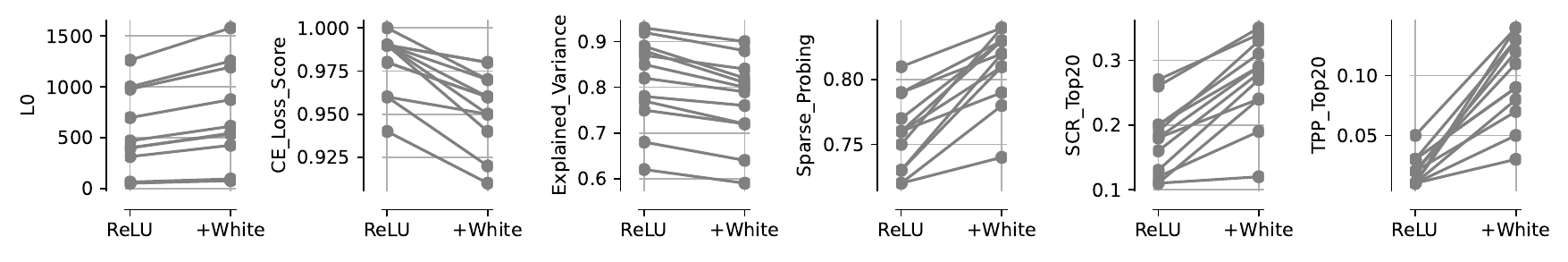}
    \caption{ReLU architecture: Each line connects paired runs before (left) and after whitening (right) averaged across all configurations (both models, all widths, all sparsity penalties). The figure illustrates significant increases in Sparse Probing, SCR, and TPP, accompanied by modest decreases in CE Loss and Explained Variance.}
    \label{fig:ReLUDiagram}
\end{figure*}

\subsection{Experimental Setup}
We evaluate the effect of PCA whitening on SAE training using SAEBench, a benchmark measuring multiple dimensions of SAE quality. Experiments were conducted on Pythia-160M (layer 8, $d {=}768$) and Gemma-2-2B (layer 12, $d {=}2304$). For each model, we trained both ReLU and Top-K SAEs with dictionary widths of $2^{14}$ and $2^{16}$, applying three sparsity levels for ReLU (0.012, 0.02, 0.06) and two target $L_0$ values for Top-K (80, 160). Each configuration was trained under two conditions—standard training and training with PCA-whitened activations.

This yields 40 total configurations (24 ReLU, 16 Top-k), with each configuration trained both with and without whitening. Training followed the open-source \textbf{\textit{dictionary\_learning}} framework \citep{karvonen2025saebench}; full configuration details are provided in Appendix~\ref{appendix:hyperparams}.

\subsection{PCA Whitening}
Before training begins, we fit a PCA Whitener to the model’s activations collected from the target layer. Activations are sampled in batches of 2048. After sufficient activations are collected to estimate the covariance matrix, we compute the whitening parameters: the mean vector \(\mu\), the whitening matrix \(W\), and the dewhitening matrix \(W^{-1}\).
 These parameters are computed once prior to training and remain fixed throughout both training and evaluation.
To compute the whitening transformation, the activation matrix \(X\) is first mean-centered:
\begin{equation}
\tilde{X} = X - \mu.
\end{equation}
We then calculate the covariance matrix of the centered activations,
\begin{equation}
\Sigma = \frac{1}{n - 1} \tilde{X}^\top \tilde{X}.
\end{equation}
Next, we perform eigendecomposition on $\Sigma$,
\begin{equation}
\Sigma = E D E^\top,
\end{equation}
where \(E\) contains the eigenvectors and \(D\) is a diagonal matrix of eigenvalues. 
The eigenvectors represent the principal axes of variation in the activation space, 
while the eigenvalues quantify the variance captured along each corresponding direction. 
The whitening transformation leverages this decomposition to decorrelate features by 
rotating the activations into the principal component basis and rescaling them to have unit variance. 
Formally, the whitening matrix is defined as
\begin{equation}
W = D^{-\frac{1}{2}} E^\top,
\end{equation}
where each diagonal element of \(D^{-\frac{1}{2}}\) is given by \(1 / \sqrt{\lambda_i + \varepsilon}\), 
and a small constant \(\varepsilon\) is added to ensure numerical stability. 

A corresponding dewhitening matrix is constructed as
\begin{equation}
W^{-1} = E D^{\frac{1}{2}},
\end{equation}
which restores the original scale and covariance structure during reconstruction. Both \(W\) and \(W^{-1}\) are stored and reused across all forward passes once fitted.

During SAE training, activations are first mean-centered and projected into the whitened space via \(W\). The encoder learns sparse representations from these decorrelated features, and the sparsity penalty is computed in the whitened space. The decoder reconstructs activations in the same space, after which the output is dewhitened using \(W^{-1}\) prior to computing the reconstruction loss. This ensures that reconstruction quality is evaluated relative to the model’s original activation distribution.

\begin{table*}[t]
\centering
\begin{tabular}{l|cc|c|c|c}
\hline
Metric & Top-k & +Whitening & $\Delta$ & \% $\Delta$ & $p$-value \\
\hline
CE Loss Score & \textbf{0.990} $\pm$ 0.002 & 0.968 $\pm$ 0.004 & -0.022 & -2.27\% & \(4.68 \cdot 10^{-4}\) \\
Explained Variance & \textbf{0.837} $\pm$ 0.025 & 0.794 $\pm$ 0.025 & -0.044 & -5.22\% & \(1.12 \cdot 10^{-4}\) \\
Sparse Probing (Top 1) & 0.754 $\pm$ 0.005 & \textbf{0.809} $\pm$ 0.008 & +0.055 & +7.30\% & \(2.62 \cdot 10^{-5}\) \\
SCR (Top 20) & 0.311 $\pm$ 0.008 & 0.304 $\pm$ 0.010 & -0.008 & -2.41\% & 0.23 \\
TPP (Top 20) & 0.141 $\pm$ 0.037 & 0.152 $\pm$ 0.031 & +0.011 & +7.96\% & 0.24 \\
\hline
\end{tabular}
\caption{Top-k architecture: SAE performance metrics averaged across all configurations (both models, all widths, all target L0s). Values shown as mean $\pm$ SEM. $\Delta$ represents the difference between whitened and standard SAE. Bold indicates significantly better performance ($p < 0.05$).}
\label{tab:topk_results1}
\end{table*}

\begin{figure*}
    \centering
    \includegraphics[width=0.99\linewidth]{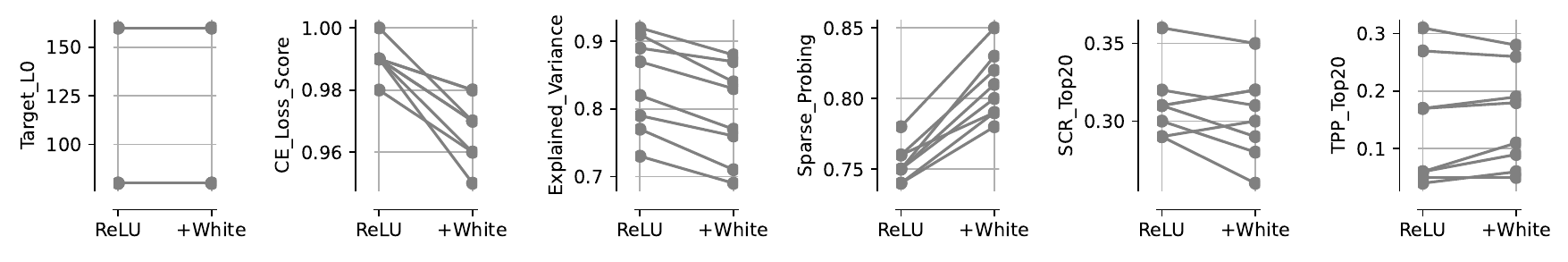}
    \caption{Top-K architecture. Each line connects paired runs before (left) and after whitening (right), averaged across all configurations (both models, widths, and target L0s). The figure shows a strong increase in Sparse Probing with no significant changes in SCR or TPP, alongside small decreases in CE Loss and Explained Variance.}
    \label{fig:topk_diagram}
\end{figure*}

\subsubsection{Evaluation} To ensure consistent preprocessing, the trained SAE is wrapped with a whitening interface during evaluation. This wrapper automatically whitens input activations before encoding and dewhitens reconstructed activations after decoding, maintaining the same transformations applied during training(\(W\) and \(W^{-1}\)).

\subsection{Results}

Our results reveal a consistent pattern: \textit{whitening significantly improves interpretability metrics while slightly reducing reconstruction quality}.

\subsubsection{Interpretability gains.} As shown in Table~\ref{tab:relu_results1} and Table~\ref{tab:topk_results1}, whitening yields substantial interpretability improvements across both ReLU and Top-K-based SAEs. For ReLU SAEs, whitening improves sparse probing accuracy by \textbf{+7.15\%}, SCR by \textbf{+54.03\%}, and TPP by \textbf{+372.00\%}, all statistically significant ($p < 0.001$). Top-K SAEs show similar gains in sparse probing (\textbf{+7.30\%}) with no significant changes in SCR or TPP. These findings indicate that whitening consistently enhances latent interpretability across architectures.

\subsubsection{Reconstruction trade-offs.} Both architecture types exhibit small yet statistically significant decreases in reconstruction metrics. For ReLU SAEs, CE Loss and explained variance decrease by \textbf{2.64\%} and \textbf{5.02\%}, respectively, while Top-K SAEs show reductions of \textbf{2.27\%} and \textbf{5.22\%}. 
While these decreases are modest compared to the substantial gains in interpretability, it is interesting to observe that the optimal sparsity–reconstruction trade-off may not coincide with peak interpretability.
Our whitening method gives equal importance to all directions in the data; however, some of those may be noise.
Thus, it may be the case that our model allocates compression budget to noisy directions.
This could be improved in future work by combining whitening with denoising \citep{chen2018sparse}.
An alternative approach for mitigating the CE drop is to replace the reconstruction term in the objective function with the final CE loss (after reconstruction, unwhitening, and the rest of the model forward pass). We leave these possible extensions for future work.

\subsubsection{Architecture-specific effects.} We observe that the benefits of whitening are more pronounced for ReLU-based architectures across all metrics. This suggests that differences in interpretability improvements across SAE variants may partly arise from implicit whitening effects or improved optimization conditioning inherent to certain architectures.


\subsubsection{Visual summary.} Figure~\ref{fig:ReLUDiagram} and Figure~\ref{fig:topk_diagram} provide a paired comparison of metric changes before and after whitening for both ReLU and Top-K architectures.

\subsection{Interpretation}

Our results challenge the conventional paradigm that optimizing for the sparsity–fidelity trade-off alone yields interpretable features. Despite mild drops in reconstruction quality, PCA-whitened SAEs produced significantly more interpretable representations. This finding is consistent with \citet{karvonen2025saebench}, who report that the Matryoshka SAE achieves the best scores on several SAEBench interpretability metrics while performing worse on the sparsity–fidelity frontier. These results support our theoretical prediction: feature formation based on data structure rather than variance leads to more interpretable features.

Among SAE variants, ReLU models benefit most from whitening. ReLU SAEs impose soft sparsity, enabling distributed representations where features of varying strength encode a concept. Whitening stabilizes these co-activation patterns, helping ReLU SAEs form more disentangled and semantically aligned features. In contrast, Top-K SAEs enforce hard sparsity, activating a fixed number of features per input regardless of concept complexity. This constraint discards weaker yet informative activations, suppressing distributed representations  needed for higher-order disentanglement. Whitening still improves the semantic alignment of active features---reflected in higher Sparse Probing---but is less effective for SCR or TPP under rigid sparsity. 

\section{Limitations and Future Work} 
Our experiments focus on middle layers (layer 8 for Pythia-160M, layer 12 for Gemma-2-2B) 
of language models at the sub-2B parameter scale. While this follows standard SAEBench 
evaluation protocols, future work should explore: (i) whitening's effects across different 
layers (early vs. late), (ii) larger model scales ($>$2B parameters), (iii) other architectures 
and modalities (vision, audio), and (iv) interactions between whitening and other training 
innovations (e.g., gated SAEs, transcoders). Additionally, investigating when the modest 
reconstruction quality drops become consequential for downstream tasks would provide 
practical guidance on when to apply whitening.

\section{Conclusion}

We have demonstrated that PCA whitening---a classical technique from sparse coding---substantially improves modern sparse autoencoder training. Through theoretical analysis, simulation, and benchmarking, whitening is found to reshape the optimization landscape, yielding significantly more interpretable features. Our results indicate that whitening should be strongly considered as a preprocessing step 
in SAE training, especially when interpretability is the primary objective.

Moreover, we find that optimizing along the sparsity--fidelity frontier alone does not necessarily yield interpretable representations. This highlights the need for a deeper
understanding of how activation geometry shapes learned features. As interpretability becomes increasingly essential for advancing scientific understanding and discovery, whitening offers a simple yet effective means to reveal the structure underlying neural
representations.
As the interpretability community continues to develop more sophisticated SAE architectures,
whitening offers a complementary preprocessing approach that can enhance any variant by
improving the underlying optimization geometry.

\section*{Acknowledgments}
We thank the SAE Bench team for providing comprehensive evaluation tools and the broader interpretability community for valuable discussions.

\bibliography{aaai2026}

\appendix

\section{Hyperparameter Configuration}
\label{appendix:hyperparams}

\begin{table}[H]
\centering
\begin{tabular}{l|l}
\hline
\textbf{Hyperparameter} & \textbf{Value} \\
\hline
Tokens processed & 500M \\
Learning rate & $5 \times 10^{-5}$ \\
Learning rate warmup (from 0) & 1{,}000 steps \\
Sparsity penalty warmup (from 0) & 5{,}000 steps \\
Learning rate decay (to 0) & Last 20\% of training \\
Dataset & The Pile \\
Batch size & 2{,}048 \\
LLM context length & 1{,}024 \\
\hline
\end{tabular}
\caption{SAE training hyperparameters.}
\label{tab:sae-hparams}
\end{table}

All Sparse Autoencoders (SAEs) were trained using the \textit{dictionary\_learning} repository, following the hyperparameter configurations specified in \textit{SAEBench} \citep{karvonen2025saebench}. Each SAE was trained on 500M tokens following this configuration. For each [layer, width, architecture] combination, we trained SAEs in a directly comparable manner, maintaining identical data and data ordering across runs. Top-$k$ SAEs were trained with target $L_0$ values of 80 and 160, as \citet{karvonen2025saebench} found that although optimal sparsity levels vary substantially across tasks, moderate $L_0$ values in the range of 50--150 offer a reasonable compromise across metrics. 

To fit the PCA whitener, we collected activation batches prior to training. 
For \textit{Pythia-160M}, we collected 10 batches, yielding an activation matrix of size $20{,}480 \times 768$. 
For \textit{Gemma-2-2B}, we collected 16 batches, resulting in an activation matrix of size $32{,}768 \times 2{,}304$. 
These matrices were used to compute the whitening transformation applied before SAE training.


\section{Simulation Details}
\label{appendix:simulation}

\subsection{Data Generation}

We generated synthetic 2D data to create a controlled environment for analyzing optimization landscapes:

\paragraph{Sparse Sources.} We sampled 10,000 points from a uniform grid over $[-1, 1]^2$, then applied a rotation and nonlinear transformation to create sparse, super-Gaussian sources $\mathbf{z} \in \mathbb{R}^{10000 \times 2}$. The sources were designed to have heavy tails and sparsity structure similar to natural latent variables in neural networks.

\paragraph{True Dictionary.} We generated a random mixing matrix $\mathbf{A} \in \mathbb{R}^{2 \times 2}$ from a Gaussian distribution, shifted to ensure well-separated dictionary columns. This represents the "ground truth" feature directions we aim to recover.

\paragraph{Observed Data.} Neural activations were generated as $\mathbf{y} = \mathbf{z}\mathbf{A}$, creating correlated observations with anisotropic variance—mimicking the structure found in real neural network activations.

\paragraph{Whitened Data.} We applied PCA whitening to obtain $\mathbf{y}_{\text{white}} = \mathbf{y}\mathbf{W}_{\text{whiten}}$, where $\mathbf{W}_{\text{whiten}} = \mathbf{V}\mathbf{D}^{-1/2}\mathbf{V}^\top$ is computed from the eigendecomposition of the covariance matrix $\mathbf{C} = \mathbf{y}^\top\mathbf{y}$. The whitening matrix was normalized to preserve overall scale.

\subsection{Landscape Computation}

To visualize the complete optimization landscape, we exhaustively evaluated two metrics across all possible 2D dictionary configurations:

\paragraph{Parameter Grid.} We discretized the dictionary space using 1,024 angles for each of the two dictionary vectors: $\theta_0, \theta_1 \in \{0, \frac{2\pi}{1024}, \frac{4\pi}{1024}, \ldots, 2\pi\}$. Each point $(\theta_0, \theta_1)$ defines a candidate dictionary:
\begin{equation}
\mathbf{W}(\theta_0, \theta_1) = \begin{bmatrix} \cos\theta_0 & \sin\theta_0 \\ \cos\theta_1 & \sin\theta_1 \end{bmatrix}
\end{equation}

\paragraph{Sparsity Metric.} For each candidate dictionary $\mathbf{W}$, we computed the inverse mean L1 norm of the reconstructed sources:
\begin{equation}
\mathcal{S}(\mathbf{W}) = \frac{1}{\mathbb{E}_{\mathbf{y}}[|\mathbf{W}^{-1}\mathbf{y}|_1]}
\end{equation}
Higher values indicate sparser reconstructions, which is desirable in sparse coding.

\paragraph{Feature Recovery Metric.} We measured how well $\mathbf{W}$ recovers the true dictionary $\mathbf{A}$ using cosine similarity:
\begin{equation}\label{eq:mcc}
\mathcal{R}(\mathbf{W}, \mathbf{A}) = \max\left(a, b\right)
\end{equation}
where
\begin{equation}
    a = \frac{|\mathbf{A}[0] \cdot \mathbf{W}[0]| + |\mathbf{A}[1] \cdot \mathbf{W}[1]|}{2}
\end{equation}
and
\begin{equation}
    b = \frac{|\mathbf{A}[0] \cdot \mathbf{W}[1]| + |\mathbf{A}[1] \cdot \mathbf{W}[0]|}{2}
\end{equation}
This metric is invariant to permutations and sign flips of dictionary columns. Values near 1 indicate accurate recovery.

\paragraph{Landscape Construction.} We computed both metrics for all $1024 \times 1024 = 1{,}048{,}576$ grid points, yielding four landscape matrices:
\begin{itemize}
\item $\mathcal{S}(\mathbf{W})$ and $\mathcal{R}(\mathbf{W}, \mathbf{A})$ for non-whitened data
\item $\mathcal{S}_{\text{white}}(\mathbf{W})$ and $\mathcal{R}_{\text{white}}(\mathbf{W}, \mathbf{A}_{\text{white}})$ for whitened data
\end{itemize}

Note that for whitened data, the ground-truth dictionary is transformed to $\mathbf{A}_{\text{white}} = \mathbf{A}\mathbf{W}_{\text{whiten}}$.

\subsection{Visualization}

The landscapes were visualized as 3D surface plots with the following design choices:

\paragraph{Figure Layout.} Each figure shows two side-by-side 3D surfaces comparing non-whitened (left) versus whitened (right) landscapes.

\paragraph{Surface Height and Color.} Two complementary views are provided:
\begin{itemize}
\item \textbf{View A}: Surface height represents feature recovery $\mathcal{R}$, colored by sparsity $\mathcal{S}$ (viridis colormap)
\item \textbf{View B}: Surface height represents sparsity $\mathcal{S}$, colored by feature recovery $\mathcal{R}$ (magma colormap)
\end{itemize}

\paragraph{Camera Angle.} All 3D plots use elevation = 40° and azimuth = 10° for consistent viewing angles.

\subsection{Key Observations}

The exhaustive landscape analysis revealed several critical differences:

\begin{enumerate}
\item \textbf{Isotropy}: Whitened landscapes are approximately rotationally symmetric, while non-whitened landscapes show strong directional biases aligned with the principal components of the data.

\item \textbf{Convexity}: The whitened landscape exhibits a smooth, bowl-like basin around the optimum, whereas the non-whitened landscape has multiple local optima and saddle points.

\item \textbf{Objective Alignment}: In whitened space, regions of high sparsity strongly overlap with regions of accurate feature recovery. In non-whitened space, these objectives can be misaligned—pursuing sparsity may lead away from the true features.

\item \textbf{Gradient Quality}: The Hessian conditioning at the optimum is significantly better after whitening, as evidenced by the smoother, less elongated basin shape.
\end{enumerate}

These geometric properties directly translate to improved optimization: gradient descent on whitened data converges faster, is less sensitive to learning rate, and finds better minima corresponding to more interpretable features. The simulation validates our theoretical analysis and motivates the use of whitening as a standard preprocessing step for SAE training.

\end{document}